\documentclass[11pt]{article}

\usepackage[]{emnlp2021}

\usepackage{times}
\usepackage{latexsym}

\usepackage[T1]{fontenc}

\usepackage[utf8]{inputenc}

\usepackage{microtype}

\usepackage{multirow}
\usepackage{booktabs}
\usepackage{amsmath}
\usepackage{amssymb}
\usepackage{graphicx}

\usepackage{booktabs, multirow, adjustbox}
\usepackage{tablefootnote}
\usepackage{xspace}
\usepackage{makecell}

\usepackage{caption}
\usepackage{subcaption}
\usepackage{tikz,pgfplots,pgfplotstable}

\usepackage{amsmath}

\usepackage{nimbusmononarrow}
\usepackage[]{microtype}

\usepackage{amsmath}
\usepackage{amsfonts}
\usepackage{pifont}
\usepackage{booktabs,multirow,adjustbox}
\usepackage{tablefootnote}

\usepackage{tikz}
\usetikzlibrary{positioning,shapes,shadows,arrows}
\usepackage{xspace,mfirstuc,tabulary}

\newcommand{\MLLM}{MLLM}
\newcommand{\MLLMs}{MLLMs}

\title{A Primer on Pretrained Multilingual Language Models}

\author{
        Sumanth Doddapaneni$^{1,3*}$  \hspace{0.2cm} Gowtham Ramesh$^{1,3}$\thanks{* The first two authors have contributed equally.} \\ \textbf{Mitesh M. Khapra}$^{1,2,3}\thanks{$\ddag$ Corresponding author: miteshk@cse.iitm.ac.in}$ \hspace{0.2cm} \textbf{Anoop Kunchukuttan}$^{3,4}$ \hspace{0.2cm} \textbf{Pratyush Kumar}$^{3,4}$
    \\ \\
    $^1$RBCDSAI,
    $^2$IIT Madras
    $^3$AI4Bharat,
    $^4$Microsoft
}

\begin{document}

\maketitle
\begin{abstract}
Multilingual Language Models (\MLLMs) such as mBERT, XLM, XLM-R, \textit{etc.} have emerged as a viable option for bringing the power of pretraining to a large number of languages. Given their success in zero-shot transfer learning, there has emerged a large body of work in (i) building bigger \MLLMs~covering a large number of languages (ii) creating exhaustive benchmarks covering a wider variety of tasks and languages for evaluating \MLLMs~ (iii) analysing the performance of \MLLMs~on monolingual, zero-shot cross-lingual and bilingual tasks (iv) understanding the universal language patterns (if any) learnt by \MLLMs~ and (v) augmenting the (often) limited capacity of \MLLMs~ to improve their performance on seen or even unseen languages. In this survey, we review the existing literature covering the above broad areas of research pertaining to \MLLMs. Based on our survey, we recommend some promising directions of future research.
\end{abstract}

\section{Introduction}

The advent of BERT \cite{devlin-etal-2019-bert} has revolutionised the field of NLP and has lead to state of the art performance on a wide variety of tasks \cite{wang-etal-2018-glue}. The recipe is to train a deep transformer based model \cite{DBLP:journals/corr/VaswaniSPUJGKP17} on large amounts of monolingual data and then fine-tune it on small amounts of task-specific data. The pretraining happens using a masked language modeling objective and essentially results in an encoder which learns good sentence representations. These pretrained sentence representations then lead to improved performance on downstream tasks when fine-tuned on even small amounts of task-specific training data \cite{devlin-etal-2019-bert}. Given its success in English NLP, this recipe has been replicated across languages leading to many language specific BERTs such as FlauBERT (French) \cite{le2020flaubert}, CamemBERT (French) \cite{martin-etal-2020-camembert}, BERTje (Dutch) \cite{devries2019bertje}, FinBERT (Finnish) \cite{ronnqvist-etal-2019-multilingual}, BERTeus (Basque) \cite{DBLP:conf/lrec/AgerriVCBSSA20}, AfriBERT (Afrikaans) \cite{ralethe-2020-adaptation}, IndicBERT (Indian languages) \cite{kakwani2020indicnlpsuite} etc. However, training such language-specific models is only feasible for a few languages which have the necessary data and computational resources.

The above situation has lead to the undesired effect of limiting recent advances in NLP to English and a few high resource languages \cite{joshi-etal-2020-state}. The question then is \textit{How do we bring the benefit of such pretrained BERT based models to a very long list of languages of interest?} One alternative, which has become popular, is to train multilingual language models (\MLLMs) such as mBERT \cite{devlin-etal-2019-bert}, XLM \cite{DBLP:conf/nips/ConneauL19}, XLM-R \cite{DBLP:conf/acl/ConneauKGCWGGOZ20}, etc. A \MLLM~ is pretrained using large amounts of unlabeled data from multiple languages with the hope that low resource languages may benefit from high resource languages due to shared vocabulary, genetic relatedness \cite{nguyen-chiang-2017-transfer} or contact relatedness \cite{goyal-etal-2020-contact}. Several such \MLLMs~ have been proposed in the past 3 years and they differ in the architecture (e.g., number of layers, parameters, etc), objective functions used for training (e.g., monolingual masked language modeling objective, translation language modeling objective, etc), data used for pretraining (Wikipedia, CommonCrawl, etc) and the number of languages involved (ranging from 12 to 100). To keep track of these rapid advances in MLLMs, as a first step, we present a survey of all existing MLLMs clearly highlighting their similarities and differences.

While training an \MLLM~ is more efficient and inclusive (covers more languages), is there a trade-off in the performance compared to a monolingual model? More specifically, \textit{for a given language is a language-specific BERT better than a \MLLM}? For example, if one is only interested in English NLP should one use English BERT or a \MLLM. The advantage of the former is that there is no capacity dilution (i.e., the entire capacity of the model is dedicated to a single language), whereas the advantage of the latter is that there is additional pretraining data from multiple (related) languages. In this work, we survey several existing studies \cite{DBLP:conf/acl/ConneauKGCWGGOZ20, DBLP:conf/rep4nlp/WuD20, DBLP:conf/lrec/AgerriVCBSSA20,DBLP:journals/corr/abs-1912-07076,ronnqvist-etal-2019-multilingual,ro2020multi2oie,DBLP:journals/corr/abs-2007-09757, DBLP:journals/corr/abs-1912-07076, wang-etal-2020-galileo, DBLP:conf/rep4nlp/WuD20} which show that the right choice depends on various factors such as model capacity, amount of pretraining data, fine-tuning mechanism and amount of task-specific training  data.

One of the main motivations of training \MLLMs~ is to enable transfer from high resource languages to low resource languages. Of particular interest, is the ability of MLLMs to facilitate zero-shot cross-lingual transfer \cite{DBLP:conf/iclr/KWMR20} from a resource rich language to a resource deprived language which does not have any task-specific training data. To evaluate such cross-lingual transfer, several benchmarks, such as XGLUE \cite{DBLP:conf/emnlp/LiangDGWGQGSJCF20}, XTREME \cite{DBLP:conf/icml/HuRSNFJ20}, XTREME-R \cite{DBLP:journals/corr/abs-2104-07412} have been proposed. We review these benchmarks which contain a wide variety of tasks such as classification, structure prediction, question answering, and cross-lingual retrieval. Using these benchmarks, several works \cite{DBLP:conf/acl/PiresSG19, DBLP:conf/emnlp/WuD19, DBLP:conf/iclr/KWMR20, DBLP:conf/acl/ArtetxeRY20, DBLP:conf/iclr/KWMR20, dufter-schutze-2020-identifying, DBLP:journals/corr/abs-2004-09205, DBLP:journals/corr/abs-2005-00633, DBLP:journals/corr/abs-2004-14218, DBLP:conf/nips/ConneauL19, DBLP:conf/emnlp/WangCGLL19,liu-etal-2019-investigating, DBLP:conf/iclr/CaoKK20,DBLP:conf/iclr/WangXXYNC20, DBLP:journals/corr/abs-2008-09112, DBLP:conf/acl/WangJBWHT20, DBLP:conf/ijcnlp/ChiDWMH20} have studied the cross-lingual effectiveness of MLLMs and have shown that such transfer depends on various factors such as amount of shared vocabulary, explicit alignment of representations across languages, size of pretraining corpora, etc. We collate the main findings of these studies in this survey. 

While the above discussion has focused on transfer learning and facilitating NLP in low resource languages, \MLLMs~could also be used for bilingual tasks. For example, \textit{could the shared representations learnt by \MLLMs~improve Machine Translation between two resource rich languages?} We survey several works \cite{DBLP:conf/nips/ConneauL19, kakwani2020indicnlpsuite, DBLP:conf/emnlp/HuangLDGSJZ19, DBLP:conf/acl/ConneauKGCWGGOZ20, DBLP:conf/emnlp/EisenschlosRCKG19, DBLP:conf/semeval/ZampieriNRAKMDP20, libovicky-etal-2020-language, jalili-sabet-etal-2020-simalign, chen-etal-2020-accurate,zenkel-etal-2020-end, dou-neubig-2021-word, imamura-sumita-2019-recycling, DBLP:journals/corr/abs-2012-15547, DBLP:journals/corr/abs-2002-06823, liu-etal-2020-multilingual-denoising, xue2021mt5} which use MLLMs for downstream bilingual tasks such as unsupervised machine translation, cross-lingual word alignment, cross-lingual QA, etc. We summarise the main findings of these studies which indicate that \MLLMs~are useful for bilingual tasks, particularly in low resource scenarios.

The surprisingly good performance of \MLLMs~in cross-lingual transfer as well as bilingual tasks motivates the hypothesis that \MLLMs~are learning universal patterns. 
However, our survey of the studies in this space indicates that there is no consensus yet. 
While representations learnt by \MLLMs~share commonalities across languages identified by different correlation analyses, these commonalities are dominantly within languages of the same family, and only in certain parts of the network (primarily middle layers). 
Also, while probing tasks such as POS tagging are able to benefit from such commonalities, harder tasks such as evaluating MT quality remain beyond the scope as yet.
Thus, though promising, \MLLMs~do not yet represent inter-lingua.

Lastly, given the effort involved in training \MLLMs~it is desirable that it is easy to extend it to new languages which weren't a part of the initial pretraining. We review existing studies which propose methods for (a) extending \MLLMs~to unseen languages, and (b) improving the capacity (and hence performance) of \MLLMs~for languages already seen during pretraining. These range from simple techniques such as fine-tuning the \MLLM~for a few epochs on the target language to using language and task specific adapters to augment the capacity of \MLLMs.

\subsection{Goals of the survey}
Summarising the above discussion, the main goal of this survey is to review existing work with a focus on the following questions:

\begin{itemize}
    \item How are different \MLLMs~built and how do they differ from each other? (Section \ref{sec:how_are_mllms_built})
    \item What are the benchmarks used for evaluating \MLLMs? (Section \ref{sec:benchmarks})
    \item For a given language, are \MLLMs~better than monolingual LMs? (Section \ref{sec:monolingual})    
    \item Do \MLLMs~ facilitate zero-shot cross-lingual transfer? (Section \ref{sec:cross-lingual})
    \item Are \MLLMs~  useful for bilingual tasks? (Section \ref{sec:bilingual})
    \item Do \MLLMs~ learn universal patterns? (Section \ref{sec:universal_patterns})
    \item How to extend \MLLMs~ to new languages? (Section \ref{sec:extend})
    \item What are the recommendations based on this survey? (Section \ref{sec:reco})
\end{itemize}

In this survey, we focus on the multilingual aspects of language models for NLU. Hence, we do not discuss related topics like monolingual LMs, pretrained models for NLG, training of large models, model compression, etc. Our survey is thus different from existing surveys on cross-lingual word embedding models \cite{10.1613/jair.1.11640}, multilingual NMT models \cite{10.1145/3406095} and pretrained language models \cite{Qiu2020PretrainedMF, Kalyan2021AMMUSA} which do not focus on pretrained multilingual language models. To the best of our knowledge, our survey is the first work that presents a comprehensive review of multilingual aspects of pretrained language models for NLU.

\section{How are \MLLMs~built?}
\label{sec:how_are_mllms_built}
The goal of \MLLMs~to is learn a model that can generate a multilingual representation of a given text. Loosely, the model should generate similar representations in a common vector space for similar sentences and words (or words in similar context) across languages.  In this section we describe the neural network architecture, objective functions, data and languages used for building \MLLMs. We also highlight the similarities and differences between existing \MLLMs.\\

\begin{table*}[ht!]
    \centering
    \small{
    \resizebox{\linewidth}{!}{
    \begin{tabular}{lcccc|cccc|cc}
    \toprule
    \textbf{Model} & \multicolumn{4}{c}{\textbf{Architecture}}&\multicolumn{4}{c}{\textbf{pretraining}}&\multicolumn{2}{c}{\textbf{Languages}}\\
        & $N$ & $k$ & $d$ & $\#Params.$ & \textit{\begin{tabular}{@{}c@{}}Objective \\ Function \end{tabular}} & \textit{Mono.} & \textit{Parallel} & \begin{tabular}{@{}c@{}c@{}}\textit{Task} \\ \textit{specific} \\ \textit{data} \end{tabular} & \#\textit{langs.} & \textit{vocab.} \\
 
    \midrule
    IndicBERT \cite{kakwani2020indicnlpsuite} & 12  & 12 & 768 & 33M & MLM &   IndicCorp & \ding{55} & \ding{55} & 12 &  200K \\ 
    Unicoder \cite{DBLP:conf/emnlp/HuangLDGSJZ19}  & 12  & 16 & 1024 & 250M &  \begin{tabular}{@{}c@{}}MLM, TLM, \\  CLWR, CLPC, CLMLM \end{tabular}& Wikipedia & \checkmark & \ding{55} & 15 &  95K \\
    XLM-15 \cite{DBLP:conf/nips/ConneauL19}  & 12  & 8 & 1024 & 250M & MLM, TLM & Wikipedia & \checkmark & \ding{55} & 15 &  95K \\
    XLM-17 \cite{DBLP:conf/nips/ConneauL19}  & 16  & 16 & 1280 & 570M & MLM &  Wikipedia & \checkmark & \ding{55} & 17 &  200K \\
    MuRIL \cite{khanuja2021muril} &  12 & 12 & 768 & 236M & MLM, TLM & \begin{tabular}{@{}c@{}}CommonCrawl \\ + Wikipedia \end{tabular} & \checkmark & \ding{55} & 17 &  197K \\ 
    VECO-small \cite{luo2021veco} & 6  & 12 & 768 & 247M & MLM, CS-MLM${^\dagger}$ & CommonCrawl  & \checkmark & \ding{55}  & 50 & 250K  \\ 
    VECO-Large \cite{luo2021veco} & 24  & 16 & 1024 & 662M & MLM, CS-MLM & CommonCrawl  & \checkmark & \ding{55}  & 50 & 250K  \\
    XLM-align \cite{DBLP:conf/acl/Chi0ZHMHW20}  & 12  & 12 & 768 & 270M & MLM, TLM, DWA &   \begin{tabular}{@{}c@{}}CommonCrawl \\ + Wikipedia \end{tabular} & \checkmark & \ding{55} & 94 &  250K \\
    InfoXLM-base \cite{chi-etal-2021-infoxlm}  & 12  & 12 & 768 & 270M & MLM, TLM, XLCO &   CommonCrawl & \checkmark & \ding{55} & 94 &  250K \\
    InfoXLM-Large \cite{chi-etal-2021-infoxlm}  & 24  & 16 & 1024 & 559M & MLM, TLM, XLCO &   CommonCrawl & \checkmark & \ding{55} & 94 &  250K \\
    XLM-100 \cite{DBLP:conf/nips/ConneauL19}  & 16  & 16 & 1280 & 570M & MLM &  Wikipedia & \ding{55} & \ding{55} & 100 &  200K \\
    XLM-R-base \cite{DBLP:conf/acl/ConneauKGCWGGOZ20}  & 12  & 12 & 768 & 270M & MLM &   CommonCrawl & \ding{55} & \ding{55} & 100 &  250K \\
    XLM-R-Large \cite{DBLP:conf/acl/ConneauKGCWGGOZ20}  & 24  & 16 & 1024 & 559M & MLM &   CommonCrawl & \ding{55} & \ding{55} & 100 &  250K \\
    X-STILTS \cite{DBLP:conf/ijcnlp/PhangCHPLVKB20} & 24  & 16 & 1024 & 559M & MLM &   CommonCrawl & \ding{55} & \checkmark & 100 &  250K \\ 
    HiCTL-base \cite{DBLP:conf/iclr/WeiW0XYL21} & 12  & 12 & 768 & 270M & MLM, TLM, HICTL & CommonCrawl & \checkmark & \ding{55} & 100 & 250K \\
    HiCTL-Large \cite{DBLP:conf/iclr/WeiW0XYL21} & 24  & 16 & 1024 & 559M & MLM, TLM, HICTL &  CommonCrawl & \checkmark & \ding{55} & 100 & 250K \\
    Ernie-M-base \cite{ouyang2021erniem} & 12  & 12 & 768 & 270M & \begin{tabular}{@{}c@{}}MLM, TLM, \\  CAMLM, BTMLM \end{tabular}& CommonCrawl & \checkmark & \ding{55} & 100 & 250K \\
    Ernie-M-Large \cite{ouyang2021erniem} & 24  & 16 & 1024 & 559M & \begin{tabular}{@{}c@{}}MLM, TLM, \\  CAMLM, BTMLM \end{tabular} & CommonCrawl & \checkmark & \ding{55} & 100 & 250K \\
    XLM-E \cite{chi2021xlme} & 12 & 12 & 768 & 279M & MLM, TLM, MRTD, TRTD & CommonCrawl & \checkmark & \ding{55} & 100 & 250k \\
    mBERT \cite{devlin-etal-2019-bert}  &  12 & 12 & 768 & 172M & MLM &  Wikipedia & \ding{55} & \ding{55} & 104 & 110K  \\
    Amber \cite{hu2021explicit} & 12  & 12 & 768 & 172M & MLM, TLM, CLWA, CLSA & Wikipedia  & \checkmark & \ding{55} & 104 & 120K  \\
    RemBERT \cite{chung2021rethinking} & 32  & 18 & 1152 & 559M${^\ddagger}$ & MLM &  \begin{tabular}{@{}c@{}}CommonCrawl \\ + Wikipedia \end{tabular}  & \ding{55} & \ding{55} & 110
    & 250K \\ 
    \toprule
    \end{tabular}
    }
    \caption{A comparison of existing Multilingual Language Models. $\dagger$ - Cross sequence MLM which is useful for NLG tasks. $\ddagger$ - For pretraining, RemBERT uses 995M parameters }

    \label{tab:mllms_comparison}
   }
\end{table*}

\subsection{Architecture}

Multilingual Language models are typically based on the transformer architecture introduced by \citet{DBLP:journals/corr/VaswaniSPUJGKP17} and then adapted for Natural Language Understanding (NLU) by \citet{devlin-etal-2019-bert} (although there are a few exceptions like \cite{DBLP:conf/emnlp/EisenschlosRCKG19} which use RNN based models). 

\paragraph{Input Layer} The input to the MLM is a sequence of tokens. The token input comes from a one-hot representation of a finite vocabulary, which is typically a subword vocabulary. This vocabulary is generally learnt from a concatenation of monolingual data from various languages using algorithms like BPE \cite{sennrich-etal-2016-neural}, wordpiece \cite{DBLP:journals/corr/WuSCLNMKCGMKSJL16} or sentencePiece \cite{DBLP:conf/emnlp/KudoR18}. To ensure reasonable representation in the vocabulary for different languages and scripts, data can be sampled using exponential weighted smoothing (discussed later) \cite{DBLP:conf/acl/ConneauKGCWGGOZ20, devlin2018bert} or separate vocabularies can be learnt for clusters of languages \cite{DBLP:conf/emnlp/ChungGTR20} partitioning the vocab size. 

\paragraph{Transformer Layers} A typical \MLLM~ comprises the encoder of the transformer network and contains a stack of $N$ layers with each layer containing $k$ attention heads followed by a feedforward neural network. For every token in the input sequence, an attention head computes an embedding using an attention weighted linear combination of the representations of all the other tokens in the sentence. The embeddings from all the attention heads are then concatenated and passed through a feedforward network to produce a $d$ dimensional embedding for each input token. As shown in Table \ref{tab:mllms_comparison}, existing \MLLMs~may differ in the choice of $N$, $k$ and $d$. Further, the parameters in each layer may be shared as in \cite{kakwani2020indicnlpsuite}.

\paragraph{Output Layer}  The outputs of the last transformer layer are typically used as contextual representations for each token, while the embedding corresponding to the \texttt{[CLS]} token is considered to be the embedding of the entire input text. Alternatively, the text embedding can also be computed via pooling operations on the token embeddings. The output layer contains simple linear transformation followed by a softmax that takes as input a token embedding from the last transformer layer and outputs a probability distribution over the tokens in the vocabulary. Note that the output layer is required only during pretraining and can be discarded during fine-tuning and task-inference - a fact that the RemBERT model uses to reduce model size \cite{DBLP:conf/iclr/ChungFTJR21}.

\subsection{Training Objective Functions}

A variety of objective functions have been proposed for training \MLLMs. These can be broadly categorized as monolingual or parallel objectives depending on the nature of training data required. We discuss and compare these objective functions in this section. 

\subsubsection{Monolingual Objectives}

The objective functions are defined on monolingual data alone. These are unsupervised/self-supervised objectives that train the model to generate multilingual representations by predicting missing tokens given the context tokens. 

\noindent{\textbf{Masked Language Model (MLM).}} This is the most standard training objective used for training most \MLLMs. Typically, other pretraining objectives are used in conjunction with the MLM objective. It is a simple extension of the unsupervised MLM objective for a single language to multiple languages by pooling together monolingual data from multiple languages. Let $x_1, x_2, \dots, x_T$ be the sequence of words in a given training example. Of these, $k$ tokens ($\approx 15\%$) are randomly selected for masking. If the $i$-th token is selected for masking then it is replaced by (i) the \texttt{[MASK]} token ($\approx$ 80\% of the time), or (ii) a random token ($\approx$ 10\% of time), or (iii) kept as it is ($\approx$ 10\% of time). The goal is to then predict these $k$ masked tokens using the remaining $T-k$ tokens. More formally, the model is trained to minimize the cross entropy loss for predicting the masked tokens. Specifically, if $u_i \in \mathbb{R}^d$ is the representation for the $i$-th masked token computed by the last layer, then the cross-entropy loss for predicting this token is computed as:\\

\begin{equation*}
    LL(i) = - \log \frac{e^{Wu_i}}{\sum_{j=1}^Ve^{Wu_j}} 
\end{equation*}
where $V$ is the size of the vocabulary, $W \in \mathbb{R}^{V \times d}$ is a parameter to be learned. The total loss is then obtained by summing over the loss of all the masked tokens. While this objective is monolingual, it surprisingly helps learn multilingual models where the encoder representations across languages are aligned without the need for any parallel corpora. The potential reasons for this surprising effectiveness of MLM for multilingual models are discussed later. 

\noindent {\textbf{Causal Language Model (CLM).}} This is the traditional language modelling objective of predicting the next word given the previous words. Unlike MLM, CLM has access to just unidirectional context. Given the success of MLM based language models for NLU applications, CLM has fallen out of favour and is currently used for pretraining NLG models where only unidirectional context is available for generation. We describe it for the sake of completeness.  Let $x_1, x_2,$ $\dots, x_T$ be the sequence of words in a given training batch. The goal then is to predict the $i$-th word given the previous $i-1$ words. Specifically, the model is trained to minimize the cross entropy loss for predicting the $i$-th word given the previous $i-1$ words.

\noindent {\textbf{Multilingual Replaced Token Detection (MRTD).}} This objective function requires the model to detect real input tokens from the corrupted multilingual sentence. Let $x_1, x_2, \dots, x_T$ be the sequence of words in a given training example. $k$ tokens are masked and a generator $G$ (usually a smaller transformer model trained with MLM objective) is used to predict these masked tokens. A discriminator $D$ is used on top of this, which takes the sentence predicted by $G$, $X^{corrupt}$ as input and then for each $x_i^{corrput}$ in $X^{corrupt}$ predicts whether it was the original token or token generated by $G$.

\subsubsection{Parallel-corpora Objectives}
\label{sub:parallel1}
These objectives require parallel corpora and are designed to explicitly force representations of similar text across languages to be close to each other in the multilingual encoder space. The objectives are either word-level (TLM, CAMLM, CLMLM, HICTL, CLSA) or sentence-level (XLCO, HICTL, CLSA). Since parallel corpus is generally much smaller than the monolingual data, the parallel objectives are used in conjunction with monolingual models. This can be done via  joint optimization of parallel and monolingual objectives, with each objective weighted appropriately. Sometimes, the initial training period may involve only monolingual objectives (XLCO, HICTL, CLSA).

\noindent{\textbf{Translation Language Model (TLM).}} In addition to monolingual data in each language, we may also have access to parallel data between some languages. \citet{DBLP:conf/nips/ConneauL19} introduced the translation language modeling objective to leverage such parallel data. Let $x_1^\mathcal{A}, x_2^\mathcal{A}, \dots, x_T^\mathcal{A}$ be the sequence of words in a language $\mathcal{A}$ and let $x_1^\mathcal{B}, x_2^\mathcal{B}, \dots, x_T^\mathcal{B}$ be the corresponding parallel sequence of words in a language $\mathcal{B}$. Both the sequences are fed as input to the MLM with a \texttt{[SEP]} token in between. Similar to MLM, a total of $k$ tokens are masked such that these tokens could either belong to the sequence in $\mathcal{A}$ or the sequence in $\mathcal{B}$. To predict a masked word in $\mathcal{A}$ the model could rely on the surrounding words in $\mathcal{A}$ or the translation in $\mathcal{B}$, thereby implicitly being forced to learn aligned representations. More specifically, if the context in $\mathcal{A}$ is not sufficient (due to masking or otherwise), then the model can use the context in $\mathcal{B}$ to predict a masked token in $\mathcal{A}$. The final objective function is the same as MLM, i.e., to minimise the cross entropy loss of the masked tokens. The only difference is that the masked tokens could belong to either language.

\noindent{\textbf{Cross-attention Masked Language Modeling (CAMLM).}}
CAMLM introduced in \citet{ouyang2021erniem} learns cross-lingual representation by predicting masked tokens in a parallel sentence pair. While predicting the masked tokens in a source sentence, the model is restricted to use semantics of the target sentence and vice versa. As opposed to TLM, where the model has access to both input sentence pairs to predict the mask tokens, in CAMLM, the model is restricted to only use the tokens in the corresponding parallel sentence to predict the masked tokens in the source sentence.

\noindent{\textbf{Cross-lingual Masked Language Modeling (CLMLM).}}
 CLMLM \cite{DBLP:conf/emnlp/HuangLDGSJZ19} is very similar to TLM objective as the masked-language-modeling is performed with cross-lingual sentences as input. The main difference is that unlike TLM, the input is constructed at document level where multiple sentences from a cross-lingual document are replaced by their translations in another language.

\noindent{\textbf{Cross-lingual Contrastive Learning (XLCO).}} \citet{chi-etal-2021-infoxlm} propose that we can leverage parallel data for training \MLLMs~ by maximizing the information content between parallel sentences. For example, let $a_i^\mathcal{A}$ be a sentence in language $\mathcal{A}$ and let $b_i^\mathcal{B}$ be its translation in language $\mathcal{B}$. In addition, let $\{b_j^\mathcal{B}\}_{j=1, j\neq i}^N$ be $N-1$ sentences in $\mathcal{B}$ which are not translations of $a_i^\mathcal{A}$.  \citet{chi-etal-2021-infoxlm} show that the information content between $a_i^\mathcal{A}$ and $b_i^\mathcal{B}$ can be maximized by minimizing the following InfoNCE \cite{oord2019representation} based loss function:

\begin{equation}
\label{eq:xlco}
    L_{XLCO} = - \log \frac{\exp(f(a_i^\mathcal{A})^\top f(b_i^\mathcal{B}))}{\sum_{j=1}^N \exp(f(a_i^\mathcal{A})^\top f(b_j^\mathcal{B})} 
\end{equation}

\noindent where $f(a)$ is the encoding of the sentence $a$ as computed by the transformer. Instead of just explicitly sampling negative sentences from $\{b_j^\mathcal{B}\}_{j=1, j\neq i}^N$, the authors use momentum \cite{he2020momentum} and mixup contrast \cite{chi-etal-2021-infoxlm} to construct harder negative samples.

\noindent{\textbf{Hierarchical Contrastive Learning (HICTL).}}
HICTL introduced in \citet{DBLP:conf/iclr/WeiW0XYL21} also uses an InfoNCE based contrastive loss (CTL) and extends it to learn both sentence level and word level cross-lingual representations. For sentence level CTL (same as Equation \eqref{eq:xlco}), instead of directly sampling from $\{b_j^\mathcal{B}\}_{j=1, j\neq i}^N$, the authors use smoothed linear interpolation \cite{bowman-etal-2016-generating, Zheng_2019_CVPR} between sentences in the embedding space to construct hard negative samples. For word level CTL, the similarity score used in the contrastive loss is calculated between the sentence representation $q$ (\texttt{[CLS]} token of a parallel sentence pair  $<a_i^\mathcal{A}$, $b_i^\mathcal{B}>$) and other words. A bag of words $\mathcal{W}$ is maintained for each parallel sentence pair input and each word in $\mathcal{W}$ is considered a positive sample while the other words in the vocabulary are considered negative. Instead of sampling negative words from the large vocabulary $V$, they construct a subset $S \subset V-W$ of negative words that are very similar to $q$ in the embedding space.

 \noindent{\textbf{Cross-lingual Sentence alignment (CLSA).}} \citet{hu2021explicit} leverage parallel data by training the cross-lingual model to align sentence representations. Given a parallel sentence ($X, $Y), the model is trained to predict the corresponding translation $Y$ for sentence $X$ from negative samples in a minibatch. Unlike mBERT which encodes two sentences and uses \texttt{[CLS]} embeddings for sentence representation, the corresponding sentence representation in AMBER \cite{hu2021explicit} is computed by averaging the word embeddings in the final layer of the \MLLM.
 
\noindent{\textbf{Translation Replaced Token Detection (TRTD).}}  Similar to the monolingual MRTD objective, \citet{chi2021xlme} leverage parallel sentences in a discriminative setup. In particular, they concatenate parallel sentences to form a single input sentence. They then use a generator $G$ to predict the masked tokens and then pass the corrupted sentence $X^{corrput}$ to a discriminator $D$ which does token level classification to discriminate between generated tokens and original tokens.

 \subsubsection{Objectives based on other parallel resources}
 \label{sub:parallel2}
 
 While parallel corpora are the most commonly used resource in the parallel objectives, other objectives use additional parallel resources like word alignments (CLWR, CLWA), cross-lingual paraphrases (CLPC), code-mixed data (ALM) and backtranslated data (BTMLM).

\noindent{\textbf{Cross-lingual word recovery (CLWR):}}
Similar to TLM, this task introduced by \citet{DBLP:conf/emnlp/HuangLDGSJZ19} aims to learn the word alignments between parallel sentences in two languages. A trainable attention matrix \cite{bahdanau2016neural} is used to represent source language word embeddings by the target language word embeddings. The cross-lingual model is then trained to reconstruct the source language word embeddings from the transformation.
 
 \noindent{\textbf{Cross-lingual paraphrase classification (CLPC).}}
\citet{DBLP:conf/emnlp/HuangLDGSJZ19} leverage parallel data by introducing a paraphrase classification objective where parallel sentences ($X, $Y) across languages are positive samples and non-parallel sentences ($X, $Z) are treated as negative samples. To make the task challenging, they train a lightweight paraphrase detection model and sample $Z$ that is very close to $X$ but is not equal to $Y$
 
  \noindent{\textbf{Alternating Language Model (ALM).}} ALM \cite{Yang_Ma_Zhang_Wu_Li_Zhou_2020} uses parallel sentences to construct code-mixed sentences and perform MLM on it. The code mixed sentences are constructed by replacing aligned phrases between source and target language.
 
 \noindent{\textbf{Denoising word alignment (DWA) and self-labeling.}} \citet{DBLP:conf/acl/Chi0ZHMHW20} leverage parallel data by learning word alignment based objective. The objective follows two alternating steps which are optimized with expectation-maximization: \\ 
 i) Self-labeling - Given a parallel sentence pair ($X, $Y) of length $n$, $m$ respectively, they first learn a doubly stochastic word alignment matrix $A$, where $A_{ij}$ gives the alignment probabiity of word $X_i$ with $Y_j$. This problem is framed as an optimal transport problem and its values are iteratively updated with Sinkhorn's algorithm \cite{Peyr2019ComputationalOT}
 \\
 ii) Denoising word alignment - Similar to TLM, some tokens in the parallel sentence pair are masked. The forward alignment probability of a masked token is calculated as follows: \\
 \begin{align}
    a_i &= {\mathrm{softmax}}(\frac{q_i^\top K}{\sqrt{d_h}}) \\
    q_i &= \mathrm{linear} ( h^*_i ) \\
    K &= \mathrm{linear} ( [h^{*}_{n+1} \dots h^{*}_{n+m}] )
\end{align}
where $i$ is the position of the masked token in the source language, $h$ is the hidden state representations from the encoder, $h^*_i$ is the query vector and $[h^{*}_{n+1} \dots h^{*}_{n+m}]$ are the key vectors represented by hidden states of target tokens,  $d_h$ is the dimension of the hidden states. The backward alignment is similarly calculated by having hidden states of source tokens as key vectors and query vectors with target sentence tokens. Given the self-labeled word alignments from previous step, the objective is to minimize the cross entropy loss between alignment probability $a_i$ and self-labeled word alignment $A_i$.
\\
  \noindent{\textbf{Cross-lingual Word alignment (CLWA).}} \citet{hu2021explicit} leverage attention mechanism in transformers to learn a word alignment based objective with parallel data. The model is trained to produce two attention matrices - source to target attention $A_{x\rightarrow y}$ which measures the similarity of source words with target words and similarly target to source attention $A_{y\rightarrow x}$. To encourage the model to align words similarly in both source and target direction, they minimise the distance between $A_{x\rightarrow y}$ and $A_{y\rightarrow x}$.
 
  \noindent{\textbf{Back Translation Masked Language Modeling (BTMLM).}} BTMLM introduced in \citet{ouyang2021erniem} attempts to overcome the limitation of unavailability of parallel corpora for learning cross-lingual representations by leveraging back translation \cite{sennrich2016improving}. It has 2 stages, wherein in the first stage, pseudo-parallel data is generated from a given monolingual sentence. In ERNIE-M \cite{ouyang2021erniem}, the pseudo-parallel sentence is generated by pretraining the model first with CAMLM and adding placeholder masks at the end of the original monolingual sentence to indicate the position and language the model needs to generate. In the second stage, tokens in the original monolingual sentence are masked and the sentence is then concatenated with the generated pseudo-parallel sentence. The model has to then predict the masked tokens.

 In summary, parallel data can be used to improve both word level and sentence level cross lingual representations. The word alignment based objectives help for zero-shot transfer on word level tasks like POS, NER, etc while the sentence level objectives are useful for tasks like cross-lingual sentence retrieval. Table \ref{tab:mllms_comparison} summarises the objective functions used by existing \MLLMs.

\subsection{Pretraining Data}

\noindent\textbf{\underline{Pretraining data.}} During pretraining \MLLMs~ use two different sources of data (a) large monolingual corpora in individual languages, and (b) parallel corpora between some languages. Existing \MLLMs~ differ in the source of monolingual corpora they use. For example, mBERT \cite{devlin-etal-2019-bert} is trained using Wikipedia whereas XLM-R \cite{DBLP:conf/acl/ConneauKGCWGGOZ20} is trained using the much larger common-crawl corpus. IndicBERT \cite{kakwani2020indicnlpsuite} on the other hand is trained on custom crawled data in Indian languages. These pretraining data-sets used by different \MLLMs~ are summarised in Table \ref{tab:mllms_comparison}. \\

\noindent\textbf{\underline{Languages.}} Some \MLLMs~ like XLM-R are  massively multilingual as they support $\sim100$ languages whereas others like IndicBERT \cite{kakwani2020indicnlpsuite} and MuRIL \cite{khanuja2021muril} 
 support a smaller set of languages. When dealing with large number of languages one needs to be careful about the imbalance between the amount of pretraining data available in different languages. For example, the number of English articles in Wikipedia and CommonCrawl is much lager than the number of Finnish or Odia articles. Similarly, there might be a difference between the amount of parallel data available between different languages. To ensure that low resource languages are not under-represented in the model, most \MLLMs~use exponentially smoothed weighting of the data while creating the pretraining data. In particular, if $m\%$ of the total pretraining data belongs to language $i$ then the probability of that language is $p_i=\frac{k}{100}$. Each $p_i$ is exponentiated by a factor $\alpha$ and the resulting values are then normalised to give a probability distribution over the languages. The pretraining data is then sampled according to this distribution. If $\alpha < 1$ then the net effect is that the high-resource languages will be under sampled and the low resource languages will be over sampled. This also ensures that the low resource languages get a reasonable share of the total vocabulary used by the model (while training the wordpiece \cite{Schuster2012JapaneseAK} or sentencepiece model \cite{kudo-richardson-2018-sentencepiece}).  Table \ref{tab:mllms_comparison} summarises the number of languages supported by different \MLLMs~ and the total vocabulary used by them. Typically, \MLLMs~ which support more languages have a larger vocabulary.

\section{What are the benchmarks used for evaluating MLLMs?}
\label{sec:benchmarks}

The most common evaluation for \MLLMs~ is cross-lingual performance on downstream tasks, \textit{i.e.}, fine-tune the model on task-specific data for a high-resource language like English and evaluate it on other languages. Some common cross-lingual benchmarks are XGLUE \cite{DBLP:conf/emnlp/LiangDGWGQGSJCF20}, XTREME \cite{DBLP:conf/icml/HuRSNFJ20}, XTREME-R \cite{DBLP:journals/corr/abs-2104-07412}. These benchmarks contain training/evaluation data for a wide variety of tasks and languages as shown in Table \ref{tab:mlm_benchmarks}. These tasks can be broadly classified into the following categories as discussed below (i) classification, (ii) structure prediction, (iii) question answering, and (iv) retrieval. \\

\noindent \textbf{Classification.} Given an input comprising of a single sentence or a pair of sentences, the task here is to classify the input into one of $k$ classes. For example, consider the task of Natural Language Inference (NLI) where the input is a pair of sentences and the output is one of 3 classes: \textit{entails, neutral, contradicts}. Some of the popular text classification datasets used for evaluating \MLLMs~are XNLI \cite{conneau2018xnli}, PAWS-X \cite{pawsx2019emnlp} , XCOPA \cite{ponti2020xcopa}, NC \cite{DBLP:conf/emnlp/LiangDGWGQGSJCF20}, QADSM \cite{DBLP:conf/emnlp/LiangDGWGQGSJCF20}, WPR \cite{DBLP:conf/emnlp/LiangDGWGQGSJCF20} and QAM \cite{DBLP:conf/emnlp/LiangDGWGQGSJCF20}.

\begin{table*}[!t]
\setlength{\tabcolsep}{3pt}
\renewcommand{\arraystretch}{1.2}
\resizebox{\linewidth}{!}{
\begin{tabular}{ccccccccccc}
\toprule
\textbf{Task} & \textbf{Corpus} & \textbf{Train} & \textbf{Dev} & \textbf{Test} & \textbf{Test Sets} & \textbf{Lang} & \textbf{Task} & \textbf{Metric} & \textbf{Domain} & \textbf{Benchmark} \\
\hline
\cline{1-2}
\hline
\multirow{7}{*}{Classification} & XNLI & 392,702 & 2,490 & 5,010 & Translations & 15 & NLI & Acc. & Misc. & XT, XTR, XG \\
 & PAWS-X & 49,401 & 2,000 & 2,000 & Translations & 7 & Paraphrase & Acc. & Wiki / Quora & XT, XTR, XG \\
 & XCOPA & 33,410+400 & 100 & 500 & Translations & 11 & Reasoning & Acc. & Misc & XTR \\
 & NC & 100k & 10k & 10k & - & 5 & Sent. Labelling & Acc. & News & XG \\
 & QADSM & 100k & 10k & 10k & - & 3 & Sent. Relevance & Acc. & Bing & XG \\
 & WPR & 100k & 10k & 10k & - & 7 & Sent. Relevance & nDCG & Bing & XG \\
 & QAM & 100k & 10k & 10k & - & 7 & Sent. Relevance & Acc. & Bing & XG \\
 \hline
\cline{1-2}
\hline
\multirow{3}{*}{Struct. Pred} & UD-POS & 21,253 & 3,974 & 47-20,436 & Ind. annot. & 37(104) & POS & F1 & Misc. & XT, XTR, XG \\
 & WikiANN-NER & 20,000 & 10,000 & 1,000-10,000 & Ind. annot. & 47(176) & NER & F1 & Wikipedia & XT, XTR \\
 & NER & 15k & 2.8k & 3.4k & - & 4 & NER & F1 & News & XG \\
 \hline
\cline{1-2}
\hline
\multirow{3}{*}{QA} & XQuAD & 87,599 & 34,736 & 1,190 & Translations & 11 & Span Extraction & F1/EM & Wikipedia & XT, XTR \\
 & MLQA &  &  & 4,517-11,590 & Translations & 7 & Span Extraction & F1/EM & Wikipedia & XT, XTR \\
 & TyDiQA-GoldP & 3,696 & 634 & 323-2,719 & Ind. annot. & 9 & Span Extraction & F1/EM & Wikipedia & XT, XTR \\
 \hline
\cline{1-2}
\hline
\multirow{4}{*}{Retrieval} & BUCC & - & - & 1,896-14,330 & - & 5 & Sent. Retrieval & F1 & Wiki / News & XT \\
 & Tatoeba & - & - & 1,000 & - & 33(122) & Sent. Retrieval & Acc. & Misc. & XT \\
 & Mewsli-X & 116,903 & 10,252 & 428-1,482 & ind. annot. & 11(50) & Lang. agn. retrieval & mAP@20 & News & XTR \\
 & LAReQA XQuAD-R & 87,599 & 10,570 & 1,190 & translations & 11 & Lang. agn. retrieval & mAP@20 & Wikipedia & XTR \\
 \bottomrule
\end{tabular}
}
\caption{Benchmarks for the all tasks for evaluation of MLLMs. Lang represents the number of languages considered from the entire pool of languages as part of the benchmark. Here XT refers to XTREME, XTR refers to XTREME-R and XG refers to XGLUE}
\label{tab:mlm_benchmarks}
\end{table*}

\noindent \textbf{Structure Prediction.} Given an input sentence, the task here is to predict a label for every word in the sequence. Two popular tasks here are Parts-Of-Speech (POS) tagging and Named Entity Recognition (NER). For NER, the datasets from WikiANN-NER \cite{d701bee1cabe492caf36340d6341e27b}, CoNLL 2002 \cite{tjong-kim-sang-2002-introduction} and CoNLL 2003 \cite{tjong-kim-sang-de-meulder-2003-introduction} shared tasks are used whereas for POS tagging the Universal Dependencies dataset \cite{11234/1-2895} is used. 

\noindent \textbf{Question Answering.} Here the task is to extract an answer span given a context and a question. The training data is typically available only in English while the evaluation sets are available in multiple languages. The datasets used for this task include  XQuAD \cite{artetxe-etal-2020-cross}, MLQA \cite{DBLP:conf/acl/LewisORRS20} and TyDiQA-GoldP \cite{tydiqa}.

\noindent \textbf{Retrieval.}  Given a sentence in a source language, the task here is to retrieve a matching sentence in the target language from a collection of sentences. The following datasets are used for this task: BUCC \cite{zweigenbaum-etal-2017-overview}, Tateoba \cite{DBLP:journals/tacl/ArtetxeS19}, Mewsli-X \cite{DBLP:journals/corr/abs-2104-07412}, LAReQA XQuAD-R \cite{roy-etal-2020-lareqa}. Of these Mewsli-X and LAReQA XQuAD-R are considered to be more challenging as they involve retrieving a matching sentence in the target language from a multilingual pool of sentences.

\section{Are \MLLMs~better than monolingual models?}
\label{sec:monolingual}
As mentioned earlier, pretrained language models such as BERT \cite{devlin-etal-2019-bert} and its variants have achieved state of the art results on many NLU tasks. The typical recipe is to first pretrain a BERT-like model on large amounts of unlabeled data and then fine-tune the model with training data for a specific task in a language $L$. Given this recipe, there are two choices for pretraining: (i) pretrain a monolingual model using monolingual data from $L$ only, or (ii) pretrain a \MLLM~ using data from multiple languages (including $L$). 

The argument in favor of the former is that, since we are pretraining a model for a specific language there is no capacity dilution (i.e., all the model capacity is being used to cater to the language of interest). The argument in favor of the latter is that there might be some benefit of using the additional pretraining data from multiple (related) languages. Existing studies show that there is no clear winner and the right choice depends on a few factors as listed below:\\

\noindent \textbf{\underline{Model capacity.}} \citet{DBLP:conf/acl/ConneauKGCWGGOZ20} argue that using a high capacity \MLLM~trained on much larger pretraining data is better than smaller capacity \MLLMs. In particular, they compare the performance of mBERT (172M parameters), XLM-R$_{base}$ (270M parameters) and XLM-R (559M parameters) with a state of the art monolingual models, i.e., BERT (335M parameters), RoBERTa (355M paramaters) \cite{Liu2019RoBERTaAR} on the following datasets: XNLI \cite{conneau2018xnli}, NER \cite{tjong-kim-sang-2002-introduction, tjong-kim-sang-de-meulder-2003-introduction}, QA \cite{DBLP:conf/acl/LewisORRS20}, MNLI \cite{N18-1101}, QNLI \cite{DBLP:conf/emnlp/WangSMHLB18}, QQP \cite{WinNT, DBLP:conf/emnlp/WangSMHLB18}, SST \cite{socher2013recursive, DBLP:conf/emnlp/WangSMHLB18}, MRPC \cite{dolan2005automatically, DBLP:conf/emnlp/WangSMHLB18} and STS-B \cite{cer2017semeval, DBLP:conf/emnlp/WangSMHLB18}. They show that, \textit{in general}, XLM-R performs better than XLM-R$_{base}$ which in turn performs better than mBERT. While none of the \MLLMs~outperform a state of the art monolingual model, XLM-R matches its performance (within 1-2\%) on most tasks.

\noindent \textbf{\underline{Amount of pretraining data.}}  \citet{DBLP:conf/acl/ConneauKGCWGGOZ20} compare two similar capacity models pretrained on different amounts of data (Wikipedia v/s CommonCrawl) and show that the model trained on larger data consistently performs better on 3 different tasks (XNLI, NER, MLQA). The model trained on larger data is also able to match the performance of state of the art monolingual models. \citet{DBLP:conf/rep4nlp/WuD20} perform an exhaustive study comparing the performance of state of the art monolingual models (trained using in-language training data) with mBERT based models (fine-tuned using in-language training data). Note that the state of the art monolingual model used in these experiments is not necessarily a pretrained BERT based model (as for many low resource languages pretrainining with smaller amounts of corpora does not really help). They consider 3 tasks and a large number of languages: NER (99 languages), POS tagging (54 languages) and Dependency Parsing (54 languages).  
Their main finding is that for the bottom 30\% of languages mBERT's performance drops significantly compared to a monolingual model. They attribute this poor performance to the inability of mBERT to learn good representations for these languages from the limited pretraining data. However, they also caution that this is not necessarily due to ``multilingual'' pretraining as a monolingual BERT trained on these low resource languages still performs poorly compared to mBERT. The performance is worse than a state of the art non-BERT based model simply because pretraining with smaller corpora in these languages is not useful (neither in a multilingual nor in a monolingual setting).  

The importance of the amount of pretraining is also emphasised in \citet{DBLP:conf/lrec/AgerriVCBSSA20} where they show that a monolingual BERT based model pretrained with larger data (216M tokens v/s 35M tokens in mBERT) outperforms mBERT on 4 tasks: topic classification, NER, POS tagging and sentiment classification (1.5 to 10 point better across these tasks). Similarly, \citet{DBLP:journals/corr/abs-1912-07076} show that pretraining a monolingual BERT from scratch for Finnish with larger data (13.5B tokens v/s 450M tokens in mBERT) outperforms mBERT on 4 tasks: NER, POS tagging, dependency parsing and news classification. Both these works partly attribute the better performance to better tokenization and vocabulary representation in the monolingual model. Similarly \citet{ronnqvist-etal-2019-multilingual} show that for four Nordic languages (Danish, Swedish, Norwegian, Finnish) the performance of mBERT is very poor as compared to its performance on high resource languages such as English and German. Further, even for English and German, the performance of mBERT is poor when compared to the corresponding monolingual models in these languages. \citet{ro2020multi2oie} report similar results for Open Information extraction where a BERT based English model outperforms mBERT. In contrast to the results presented so far, \citet{DBLP:journals/corr/abs-2007-09757} show that across 7 different tasks and many different experimental setups, on average mBERT performs better than a monolingual Portuguese model trained on much larger data (4.8GB/992M tokens v/s ~2GB in mBERT). 

In summary, based on existing studies, there is no clear answer to the following question: \textit{``At what size of monolingual corpora does the advantage of training a multilingual model disappear?''}. Also note that most of these studies use mBERT and hence more experiments involving XLM-R$_{base}$, XLM-R$_{large}$ and other recent \MLLMs~are required to draw more concrete conclusions. 

\noindent\textbf{\underline{Tokenization}} Apart from the pretraining corpora size, \citet{DBLP:conf/acl/RustPVRG20} show that the monolingual models perform better than multilingual models due to their language specific tokenizer. 
To decouple the two factors, \textit{viz.}, the pretraining corpora size and tokenizer, they perform two experiments across 9 diverse typologically diverse languages and 5 tasks (NER, QA, Sentiment analysis, Dependency parsing, POS tagging). In the first experiment, they train two monolingual models using the same pretraining data but with two different tokenizers, one being a language specific tokenizer and the other being the mBERT tokenizer. In the second experiment, they retrain embedding layer of mBERT (the other layer weights are frozen) while using the two tokenisers mentioned above (monolingual tokenizer and mbert tokenizer). In both the experiments, in 38/48 combinations of model, task and language, they find that models which use mononlingual tokenizers perform much better than models which use the mBERT tokenizer. They further show that the better performance of monolingual tokenizers over mBERT tokenizer can be attributed to lower fertility and higher proportion of continued words (i.e., words that are tokenized to at least two sub tokens).\\

\noindent \textbf{\underline{Joint v/s individual fine-tuning.}} Once a model is pretrained, there are two choices for fine-tuning it (i) fine-tune an \textit{individual} model for each language using the training data for that language or (ii) fine-tune a \textit{joint} model by combining the training data available in all the languages. For example,  \citet{DBLP:journals/corr/abs-1912-07076} consider a scenario where NER training data is available in 4 languages. They show that a joint model fine-tuned using the training data in all the 4 languages matches the performance of monolingual models individually fine-tuned for each of these languages. The performance drop (if any) is so small that it does not offset the advantage of deploying/maintaining a single joint model as opposed to 4 individual models. \citet{DBLP:journals/corr/abs-1912-01389} report similar results for NER and strongly advocate a single joint model. Lastly, \citet{wang-etal-2020-galileo} report that for the task of identifying offensive language in social media posts, a jointly trained model outperforms individually trained models on 4 out of the 5 languages considered. However, more careful analysis involving a diverse set of languages and tasks is required to conclusively say if a jointly fine-tuned \MLLM~is preferable over individually fine-tuned language specific models. \\

\noindent \textbf{\underline{Amount of task-specific training data.}} Typically, there is some correlation between the amount of task-specific training data and the amount of pretraining data available in a language. For example, a low resource language would have smaller amounts of training data as well as pretraining data. Hence, intuitively, one would expect that in such cases of extreme scarcity, a multilingual model would perform better. However, \citet{DBLP:conf/rep4nlp/WuD20} show that this is not the case. In particular, they show that for the task of NER for languages with only 100 labeled sentences, a monolingual model outperforms an mBERT based model. The reason for this could be a mix of poor tokenization, lower vocabulary share and poor representation learning from limited pretraining data. We believe that more experiments which carefully ablate the amount of training data across different tasks and languages would help us better understand the utility of \MLLMs.\\

\noindent \textbf{Summary}: Based on existing studies it is not clear whether \MLLMs~are always better than monolingual models. We recommend a more systematic study where the above parameters are carefully ablated for a wider range of tasks and languages. 

\section{Do \MLLMs~ facilitate zero-shot cross-lingual transfer?}
\label{sec:cross-lingual}

In the context of \MLLMs, the standard zero-shot cross-lingual transfer from a source language to a target language involves the following steps: (i) pretrain a \MLLM~using training data from multiple languages (including the source and target language) (ii) fine-tune the source model on task-specific training data available in the source language (iii) evaluate the fine-tuned model on test data from the target language. 
In this section, we summarise existing studies on using \MLLMs~for cross-lingual transfer and highlight some factors which could influence their performance in such a setup.\\

\noindent \textbf{\underline{Shared vocabulary.}} Before training a \MLLM, the sentences from all languages are first tokenized by jointly training a WordPiece model \cite{Schuster2012JapaneseAK} or a SentencePiece model \cite{kudo-richardson-2018-sentencepiece}. The basic idea is to tokenize each word into high frequency subwords. The vocabulary used by the model is then a union of such subwords identified across all languages. This joint tokenization ensures that the subwords are shared across many similar (or even distant) languages. For example, the token `es' in the vocabulary would be shared by English, French, German, Spanish, \textit{etc}. It is thus possible that if a subword which appears in the testset of the target language is also present in the training set of the source language then some model performance will be transferred through this shared vocabulary. Several studies examine the role of this shared vocabulary in enabling cross-lingual transfer as discussed below.

\citet{DBLP:conf/acl/PiresSG19} and \citet{DBLP:conf/emnlp/WuD19} show that there is strong positive correlation between cross-lingual zero-shot transfer performance and the amount of shared vocabulary between the source and target language. The results are consistent across 5 different tasks (MLDoc \cite{SCHWENK18.658}, XNLI \cite{conneau2018xnli}, NER \cite{tjong-kim-sang-2002-introduction, tjong-kim-sang-de-meulder-2003-introduction}, POS \cite{11234/1-2895}, Dependency parsing \cite{11234/1-2895} ) and 5-39 different languages (the number of languages varies across tasks). However, \citet{DBLP:conf/iclr/KWMR20} present contradictory results and show that the performance drop is negligible even when the word overlap is reduced to zero synthetically (by shifting the unicode of each English character by a large constant thereby replacing it by a completely different character). On similar lines \citet{DBLP:conf/acl/ArtetxeRY20} show that cross-lingual transfer can happen even when the vocabulary is not shared, by appropriately fine-tuning all layers of the transformer except for the input embedding layer.\\

\noindent \textbf{\underline{Architecture of the \MLLM.}} The model capacity of a \MLLM~depends on the number of layers, number of attention heads per layer and the size of the hidden representations. \citet{DBLP:conf/iclr/KWMR20} show that the performance of cross-lingual transfer depends on the depth of the network and not so much on the number of attention heads. In particular, they show that decent transfer can be achieved even with a single headed attention. Similarly, the total number of parameters in the model are not as important as the number of layers in determining the performance of cross-lingual transfer. In fact, \citet{dufter-schutze-2020-identifying} argue that mBERT's multilinguality is due to its limited number of parameters which forces it to exploit common structures to align representations across languages. Another important architectural choice is the context window for self attention, i.e., the number of tokens fed as input to the \MLLM~during training.  \citet{DBLP:journals/corr/abs-2004-09205} show that using smaller context windows is useful if the pretraining data is small (around 200K sentences) but when large amounts of pretraining data is available then it is beneficial to use longer context windows. \\

\noindent \textbf{\underline{Size of pretraining corpora.}} \citet{DBLP:journals/corr/abs-2004-09205} show that cross-lingual transfer is better when mBERT is pretrained on larger corpora (1000K sentences per language) as opposed to smaller corpora (200K sentences per language). \citet{DBLP:journals/corr/abs-2005-00633} show that for higher level tasks such as XNLI and XQuAD, the performance of zero-shot transfer has a strong correlation with the amount of data in the target language used for pretraining the \MLLM. For lower level tasks such as POS, dependency parsing and NER also there is a positive correlation but not as high as that for XNLI and XQuAD.\\

\noindent \textbf{\underline{Fine-tuning strategies.}} \citet{DBLP:journals/corr/abs-2004-14218} argue that when a \MLLM~is fine-tuned its parameters change which weakens its cross-lingual ability as some of the alignments learnt during pretraining are lost. They motivate this by showing that the cross-lingual retrieval performance of a \MLLM~drops drastically when it is fine-tuned for the task of POS tagging. To avoid this, they suggest  using a continual learning framework for fine-tuning so that the model does not forget the original task (masked language modeling) that it was trained on. Using such a fine-tuning strategy they report better results on cross-lingual POS tagging, NER and sentence retrieval.\\

\noindent \textbf{\underline{Using bitext.}} While \MLLMs~show good transfer without explicitly being trained with any cross-lingual signals, it stands to reason that explicitly providing such signals during training should improve the performance. Indeed, XLM-R \cite{DBLP:conf/nips/ConneauL19} and InfoXLM \cite{chi-etal-2021-infoxlm} show that using parallel corpora with the TLM objective gives improved performance. If parallel corpus is not available then \citet{dufter-schutze-2020-identifying} suggest that it is better to train \MLLMs~with comparable corpora (say Wikipedia or CommonCrawl) than using corpora from different sources across languages. ~\\

\noindent \textbf{\underline{Explicit alignment of representations.}} 
Some works \cite{DBLP:conf/emnlp/WangCGLL19,liu-etal-2019-investigating} compare the performance of zero-shot cross-lingual transfer using (i) representations learned by \MLLMs~which are implicitly aligned and (ii) representations from monolingual models which are then explicitly aligned post-hoc using some bitext. They observe that the latter leads to better performance. Taking cue from this, \citet{DBLP:conf/iclr/CaoKK20,DBLP:conf/iclr/WangXXYNC20} propose a method to explicitly align the representations of aligned word pairs across languages while training mBERT. This is achieved by adding a loss function which minimises the Euclidean distance between the embeddings of aligned words. \citet{DBLP:journals/corr/abs-2008-09112} also report similar results by explicitly aligning the representations of aligned word pairs and further normalising the vector spaces (i.e., ensuring that the representations across all languages have zero mean and unit variance). \\

\noindent\textbf{\underline{Knowledge Distillation.}}
Both \citet{DBLP:conf/acl/WangJBWHT20} and \citet{DBLP:conf/ijcnlp/ChiDWMH20} argue that due to limited model capacity \MLLMs~cannot capture all the nuances of multiple languages as compared to a pretrained monolingual model which caters to only one language. They show that the cross-lingual performance of a \MLLM~can be improved by distilling knowledge from a monolingual model. The above works apply knowledge distillation on the task-specific setting, \textit{i.e.}, the teacher LMs are first fine-tuned on a specific task and then this knowledge is distilled to a student LM. \citet{Khanuja2021MergeDistillMP} focus on knowledge distillation in a task-agnostic setting. They distill knowledge from multiple multilingual teacher LMs (with overlapping languages) and show positive transfer from strong teachers for low resourced languages not covered by the teacher models.

\noindent\textbf{\underline{Source language used for fine-tuning.}} English is currently the most widely used source language for evaluating the cross-lingual transfer performance of \MLLMs. This could be due to easy availability of English datasets and also because popular multilingual benchmarks like XGLUE and XTREME use English as the default source language. To evaluate the effectiveness of this default choice and compare with other high resource alternatives, \citet{lin-etal-2019-choosing} formulate the task of choosing the best source as a learning to rank problem. Using handcrafted features (dataset size, word/subword overlap, genetic distance between languages, \textit{etc.}), they train Gradient-Boosting Decision Tree \cite{DBLP:conf/nips/KeMFWCMYL17} to predict the source language which would be most suited for cross-lingual transfer. They show that the source languages predicted by their model outperform other baselines on Machine Translation and give comparable results for POS tagging. Similarly, \citet{DBLP:journals/corr/abs-2106-16171} try to find the best source language when the set of target languages is large and unknown beforehand. They study mBERT and mT5 on classfication (XNLI, PAWS-X) and QA tasks (XQuAD, TyDi QA) and show that German and Russian are often more effective source languages than English. They also find that zero-shot transfer can often be improved by fine-tuning on English datasets which are machine translated to better source languages.\\

\noindent\textbf{\underline{Complexity of the task.}} As outlined in section \ref{sec:benchmarks}, the tasks used for evaluating \MLLMs~are of different complexity (ranging from binary classification to Question Answering). \citet{DBLP:journals/corr/abs-2104-07412} show that much of the progress on zero-shot transfer on existing benchmarks has not been uniform with most of the gains coming from cross-lingual retrieval tasks. Here again, the gains are mainly due to fine-tuning on other tasks and pretraining with parallel data. The progress of cross-lingual QA datasets (such as MLQA) is very minimal and the overall scores on the QA task are much less when compared to monolingual QA in English. Similarly, on structure prediction tasks like NER and POS tagging, there isn't much improvement in the performance going from some of the earlier models like XLM-R \cite{DBLP:conf/acl/ConneauKGCWGGOZ20} to some of the more recent models like VECO \cite{luo2021veco}. They recommend that (i) some of the easier tasks such as BUCC and PAWS-X should be dropped from these evaluation benchmarks and (ii) more complex tasks such as cross-lingual retrival from a mixed multilingual pool should be added (e.g., LAReQA \cite{roy-etal-2020-lareqa} and Mewsli-X \cite{DBLP:journals/corr/abs-2104-07412}).\\

\begin{figure*}[t!]
    \centering
    \includegraphics[width=\textwidth]{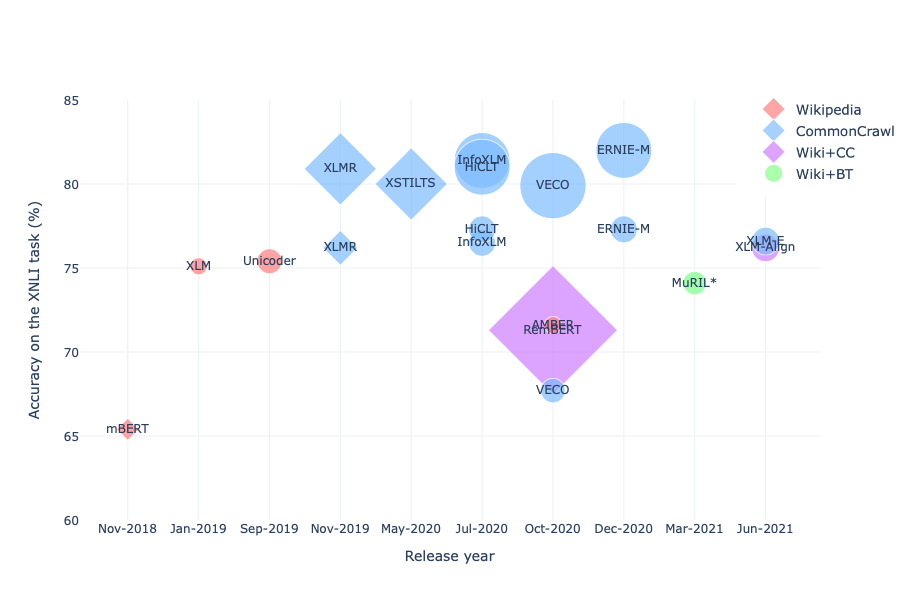}
    \caption{Comparison of the performance of different \MLLMs~on the XNLI task. We use XNLI as this task is used by almost all the \MLLMs. The size of each entry is proportional to the number of parameters in the model. Entries in circle use parallel data for tasks described in \ref{sub:parallel1} and \ref{sub:parallel2}. Entries in diamond use only monolingual data. MuRIL reports XLNI scores for only Indian Languages.}
    \label{fig:mllm_trends}
\end{figure*}

\noindent \textbf{Summary:} While not very conclusive, existing studies show that there is some evidence that the performance of \MLLMs~ on zero-shot cross-lingual transfer is generally better when (i) the source and target languages share some vocabulary (ii) there is some similarity between the source and target languages (iii) the \MLLM~ uses a deeper architecture (iv) enough pretraining data is available in the target languages (v) a continual learning (\textit{learning-without-forgetting}) framework is used (vi) the representations are explicitly aligned using bitext and appropriate loss functions and (vii) the complexity of the task is less. Note that in all the above cases, cross-lingual transfer using \MLLMs~performs much worse than using in-language supervision (as expected). Figure \ref{fig:mllm_trends} shows the zero-shot scores of models on the XNLI benchmark. Further, in most cases it performs worse than a \textit{translate-train}\footnote{\textbf{translate-train}: The training data available in one language is translated to the language of interest using a MT system. This (noisy) data is then used for training a model in the target language.} or a \textit{translate-test}\footnote{\textbf{translate-test}: The training data available in one (high resource) source language is used to train a model in that language. The test data from the language of interest is then translated to the source language using a MT system.} baseline. 

\section{Are \MLLMs~ useful for bilingual tasks?}
\label{sec:bilingual}

This survey has so far looked at the utility of \MLLMs~for cross-lingual tasks, where the multilingual capabilities of the \MLLMs~help in transfer learning and building universal models. Recent work has also explored the utility of \MLLMs~for bilingual tasks like word-alignment, sentence-retrieval, etc. This section analyzes the role of \MLLMs~for such bilingual tasks. 

\subsection{Word Alignment}

Recent work has shown that \MLLMs, which are trained on monolingual corpora only, can be used to learn high-quality word alignments in parallel sentences \cite{libovicky-etal-2020-language,jalili-sabet-etal-2020-simalign}. This presents a promising unsupervised alternative to statistical aligners \cite{brown-etal-1993-mathematics,och-ney-2003-systematic,ostling2016efficient,dyer-etal-2013-simple} and neural MT based aligners \cite{chen-etal-2020-accurate,zenkel-etal-2020-end}, both of which are trained on parallel corpora. Finding the best word alignment can be framed as a maximum-weight maximal matching problem in the weighted bipartite graph induced by the distance between word embeddings. Greedy, iterative and optimal transport based solutions to the problem have been proposed. The iterative solutions seem to perform the best with good alignment speed as well. Using contextual embeddings from some intermediate layers gives better word alignment than the top encoder layer. Unlike statistical aligners, these \MLLM~based aligners are inherently multilingual. They also significantly outperform aligners based on static word embeddings like FastText\footnote{https://github.com/facebookresearch/fastText}. The \MLLM~aligners can be significantly improved by aligning on parallel corpora and/or word-aligned data; fine-tuning also preserves the multilingual word-alignment \cite{dou-neubig-2021-word}. Word alignment is useful for transfer in downstream tasks using the translate-test paradigm which needs projection of spans between parallel text (e.g., question answering, POS, NER, etc).

\subsection{Cross-lingual Sentence Retrieval}

Since \MLLMs~represent sentences from different languages in a common embedding space, they can be used to find the translation of a sentence in another language using nearest-neighbour search. The sentence embedding can be obtained from the \texttt{[CLS]} token or appropriate pooling operations on the token embeddings. However, the encoder representations across different languages are not aligned well enough for high quality sentence-retrieval \cite{libovicky-etal-2020-language}. Unlike word alignment, embeddings learnt from monolingual corpora alone are not sufficient for cross-lingual sentence retrieval given the large search space.  These shortcomings can be overcome by centering of sentence embeddings \cite{libovicky-etal-2020-language} and/or fine-tuning the \MLLMs~ on parallel corpora   with contrastive/margin-based objectives \cite{DBLP:journals/corr/abs-2007-01852,DBLP:journals/corr/abs-2104-07412} and result in high-quality sentence retrieval. 

\subsection{Machine Translation}

NMT models are typically encoder-decoder models with attention trained using parallel corpora. It has been show that using BERT for initializing the encoder/decoder \cite{DBLP:conf/nips/ConneauL19,imamura-sumita-2019-recycling,DBLP:journals/corr/abs-2012-15547} or extracting features \cite{DBLP:journals/corr/abs-2002-06823} can help MT by pretraining the model with linguistic information. In particular, \citet{DBLP:conf/nips/ConneauL19} show that initialization of the encoder/decoder with a pretrained model can help in data-scarce scenarios like unsupervised NMT and low-resource NMT, and act as a substitute for backtranslation in supervised NMT scenarios. Subsequent research has drawn inspiration from the success of \MLLMs~and has shown the utility of denoising pretraining strategies specifically for sequence to sequence models \cite{liu-etal-2020-multilingual-denoising,xue2021mt5}.

To summarize, \MLLMs~ are useful for some bilingual tasks, particularly in low-resource scenarios and fine-tuning with parallel data provides added benefits. 

\section{Do \MLLMs~ learn universal patterns?}
\label{sec:universal_patterns}
The success of language models such as BERT has led to a large body of research in understanding how such language models work, what information they learn, and how such information is represented in the models. 
Many of these questions studied in `BERTology' are also relevant to multilingual language models (\MLLMs) given the similarity in the neural architectures of these networks.  
But one question relates specifically to the analysis of MLLMs - Do these models learn  and represent patterns which generalise across languages?
Such an expectation is an old one going back to the ``universals of language'' proposed in 1966 \cite{greenberg1966language} and has been studied at different times.
For instance, during the onset of word embeddings, it was shown that the embeddings learnt across languages can be aligned effectively \cite{mikolov2013exploiting}.
This expectation is renewed due to MLLMs demonstrating surprisingly high cross-lingual transfer as summarised in \S \ref{sec:cross-lingual}.

\noindent \textbf{Inference on parallel text.} 
Different works approach this question of universal patterns in different ways. 
One set of methods analyse the inference on MLLMs of parallel text in different languages. 
With such parallel text, the intermediate representations on the MLLM can be compared to identify alignment, quantified with different mathematical techniques such as Canonical Correlation Analysis (CCA) and Centered Kernel Alignment (CKA). 
CCA analysis for mBERT showed that the model does not project the representations of different languages on to a shared space - a trend that is stronger towards the later layers of the network \cite{singh2019bert}.
Further, the correlation of the representations of the languages mirrored language evolution, in particular phylogenetic trees discovered by linguists.
On the other hand, there exist symmetries in the representations of multiple languages as evidenced by isomorphic embedding spaces  \cite{conneau-etal-2020-emerging}.
The argument in support for the existence of such symmetries is that monolingual BERT models exhibit high degrees of CKA similarity  \cite{conneau-etal-2020-emerging}.
Another related technique to find common representations across languages is machine translation. 
Given a sentence in a source language and a few candidate sentences in a target language, can we find the correct translation by identifying the nearest neighbour in the representation space?
It is found that such translation is sensitively dependent on the layer from which the representation is learnt - peaking in the middle layers of between 5 and 8 with accuracy over 75\% for related language pairs such as English-German and Hindi-Urdu  \cite{DBLP:conf/acl/PiresSG19}.
One may conclude that the very large neural capacity in MLLMs leads to multilingual representations that have language-neutral and language-specific components. 
The language-neutral components are rich enough to align word embeddings and also retrieve similar sentences, but are not rich enough to solve complex tasks such as MT quality evaluation  \cite{libovicky2019language}.

\noindent \textbf{Probing tasks.}
Another approach to study universality is by `probing tasks' on the representations learnt at different layers.
For instance, consistent dependency trees can be learnt from the representations of intermediate layers indicating syntactic abstractions in mBERT \cite{DBLP:conf/emnlp/LimisiewiczMR20, DBLP:conf/acl/ChiHM20}.
However, the dependency trees were more accurate for Subject-Verb-Object (SVO) languages (such as English, French, Indonesian) than SOV languages (such as Turkish, Korean, and Japanese). 
This disparity between SOV and SVO languages is also observed for part-of-speech tagging \cite{DBLP:conf/acl/PiresSG19}.
Each layer has different specialisations and it is therefore useful to combine information from different layers for best results, instead of selecting a single layer based on the best overall performance as demonstrated for Dutch on a range of NLU tasks  \cite{DBLP:conf/emnlp/VriesCN20}.
In the same work, a comparison with a monolingual Dutch model revealed that a multilingual model has more informative representations for POS tagging in earlier layers.

\noindent \textbf{Controlled Ablations.}
Another set of results control for the several confounding factors which need to be controlled or ablated to check the hypothesis that MLLMs learn language-independent representations.
The first such confounding factor is the joint script between many of the high resource languages. 
This was identified not to be a sensitive factor by demonstrating that transfer between representations also occur between languages that do not share the same script, such as Urdu written in the Arabic script and Hindi written in Devanagari script  \cite{DBLP:conf/acl/PiresSG19}.
An important component of the model is the input tokenization.
There is a strong bias to learn language-independent representations when using sub-word tokenization rather than word-level or character-level \cite{singh2019bert}.
Another variable of interest is the pretraining objective. 
Models such as LASER and XLM which are trained on cross-lingual objectives retain language-neutral features in the higher layers better than mBERT and XLM-R which are only trained on monolingual objectives \cite{choenni2020does}.

In summary, there is no consensus yet on MLLMs learning universal patterns. 
There is clear evidence that MLLMs learn embeddings which have high overlap across languages, primarily between those of the same family. 
These common representations seem to be clearest in the middle layers, after which the network specialises for different languages as modelled in the pretraining objectives. 
These common representations can be probed to accurately perform supervised NLU tasks such as POS tagging, dependency parsing, in some cases with zero-shot transfer. 
However, more complex tasks such as MT quality evaluation  \cite{libovicky2019language} or language generation  \cite{ronnqvist-etal-2019-multilingual} remain outside the realm of these models currently, keeping the debate on universal patterns incomplete.

\section{How to extend MLMs to new languages}
\label{sec:extend}
Despite their success in zero-shot cross-lingual transfer, \MLLMs~suffer from the \textit{curse of multilinguality} which leads to capacity dilution. This limited capacity is an issue for (i) high resource languages as the performance of \MLLMs~for such languages is typically lower than corresponding monolingual models (ii) low resource languages where the performance is even poorer and finally (iii) languages which are unseen in training (the last point is obvious but needs to be stated nonetheless). Given this situation, an obvious question to ask is how to enhance the capacity of \MLLMs~such that it benefits languages already seen during training (high resource or low resource) as well as languages which were unseen during training. The solutions proposed in the literature to address this question can be broadly classified into four categories as discussed below. 

\noindent\textbf{Fine-tuning on the target language.} Here the assumption is that we only care about the performance on a single target language at test time. To ensure that this language gets an increased share of the model capacity we can simply fine-tune the pretrained \MLLM~using monolingual data in this language. This is akin to fine-tuning a \MLLM~ (or even a monolingual LM) for a downstream task. \citet{DBLP:conf/emnlp/PfeifferVGR20} show that such target language adaptation prior to task specific fine-tuning using source language data leads to improved performance over the standard cross-lingual transfer setting. In particular, it does not result in catastrophic
forgetting of the multilingual knowledge learned during pretraining which enables cross-lingual transfer. The disadvantage of course is that the model is now specific to the given target language and may not be suitable for other languages. Further, this method does not address the fundamental limitation in the  model capacity and still hinders adaptation to low resource and unseen languages. 

\noindent\textbf{Augmenting vocabulary.} A simple but effective way of extending a \MLLM~ to a new language which was unseen during training is to augment the vocabulary of the model with new tokens corresponding to the target language. This would in turn lead to additional parameters getting created in the input (embedding) layer and the decoder (output) layer. The pretrained \MLLM~ can then be further trained using monolingual data from the target language so that the newly introduced parameters get trained. \citet{DBLP:conf/emnlp/WangK0R20} show that such an approach is not only effective for unseen languages but also benefits languages that are already present in the \MLLM. The increase in the performance on languages already seen during training is surprising and can be attributed to (i) increased representation of this language in the vocabulary (ii) focused fine-tuning on the target language and (iii) increased monolingual data for the target language. 
Some studies have also considered unseen languages which share vocabulary with already seen languages. For example,  \citet{muller2020multilingual} show that for Naribazi (North African Arabic dialect) which is written using Latin script and has a lot of code mixing with French, simply fine-tuning the BERT with very few sentences from Naribazi (around 50000 sentences) leads to reasonable zero-shot transfer from French. For languages having an unseen script, \citet{DBLP:journals/corr/abs-2010-12858} suggest that such languages should be transliterated
to a script which is used by a related language seen
during pretraining. 

\noindent\textbf{Exploiting Latent Semantics in the embedding matrix.} \citet{DBLP:journals/corr/abs-2012-15562} argue that lexically overlapping tokens play an important role in cross-lingual transfer. However, for a new language with an unseen script such transfer is not possible. To leverage the multilingual knowledge learned in the embedding matrix, they factorise the embedding matrix ($\mathbb{R}^{|V| \times d }$) into lower dimensional word embeddings ($\mathbf{F} \in \mathbb{R^{|V| \times d_1}}$) and $C$ shared up-projection matrices ($\mathbf{G_1, G_2, G_C} \in \mathbb{R}^{d_1 \times d}$). The matrix  $\mathbf{F}$ encodes token specific information and the up-projection matrices encode general cross-lingual information. Each token is associated with one of the $C$ up-projection matrices. For an unseen language $T$ with an unseen script, they then learn an embedding matrix ($\mathbf{F'} \in \mathbb{R^{|V_T| \times d_1}})$  and an assignment for each token in that language to one of the $C$ up-projection matrices. This allows the token to leverage the multilingual knowledge already encoded in the pretrained \MLLM~via the up-projection matrices.

\noindent\textbf{Using Adapters.} Another popular way to increase model capacity is to use \textit{adapters} \cite{NIPS2017_e7b24b11,pmlr-v97-houlsby19a} which essentially introduce a small number of additional parameters for every language and/or task, thereby augmenting the limited capacity of \MLLMs. Several studies have shown the effectiveness of using such adapters in \MLLMs~\cite{DBLP:journals/corr/abs-2004-14327, DBLP:journals/corr/abs-2012-06460,DBLP:conf/emnlp/PfeifferVGR20,DBLP:journals/corr/abs-2101-03289}. We explain the overall idea by referring to the work of \citet{DBLP:conf/emnlp/PfeifferVGR20}. An adapter can be added at every layer of the transformer. Let $h$ be the hidden size of the Transformer and $d$ be the dimension of the adapter. An adapter layer simply contains a down-projection ($D: \mathbb{R}^h \rightarrow \mathbb{R}^d$) followed by a ReLU activation and an up-projection (($U: \mathbb{R}^d \rightarrow \mathbb{R}^h$). Such an adapter layer thus introduces $2 * d * h$ additional parameters for every language. During adaptation, the rest of the parameters of the \MLLM~ are kept fixed and the adapter parameters are trained using unlabelled data from the target language using the MLM objective.

Further, \citet{DBLP:conf/emnlp/PfeifferVGR20} propose that during task-specific fine-tuning, the language adapter of the source language should be used whereas during zero-shot cross-lingual transfer, at test time, the source language adapter should be replaced by the target language adapter. However, this requires that the underlying \MLLM~ does not change during fine-tuning. Hence, they add a task-specific adapter layer on top of a language specific adapter layer. During task-specific fine-tuning, only the parameters of this adapter layer are trained, leaving the rest of the \MLLM~unchanged.  \citet{DBLP:journals/corr/abs-2012-06460} use orthogonality constraints to ensure that the language and task specific adapters learn complementary information.
Lastly, to account for unseen languages whose vocabulary is not seen during training, they add invertible adapters at the input layer. The function of these adapaters is to learn token level characteristics. This adapter is also trained with the MLM objective using the unlabelled monolingual data in the target language.

In the context of the above discussion on adapters, we would like to point the readers to AdapterHub.ml \footnote{https://github.com/Adapter-Hub/adapter-transformers} which is a useful repository containing all recent adapter architectures \cite{DBLP:journals/corr/abs-2007-07779}.

\section{Recommendations}
\label{sec:reco}
Based on our review of the literature on \MLLMs~we make the following recommendations: \\

\noindent \textbf{Ablation Studies.} 
The design of deep neural models involves various parameters which are often optimized only by exhaustive ablation studies. 
In the case of \MLLMs, the axes of ablation belong to three sets - architectural parameters, pretraining objectives, and subset of languages chosen.
Given the number of options for each of these sets, an exhaustive ablation study would be prohibitively expensive.
However, in the absence of such a study some questions remain open: For instance, what subset of languages should one choose for training a multilingual model?
How should the architecture be shaped as we change the number of languages?
One research direction is to design controlled and scaled-down ablation studies where a broader set of parameters can be evaluated and generalized guidelines can be derived. \\

\noindent \textbf{Zero-Shot Evaluation.} 
The primary promise of \MLLMs~remains cross-lingual performance, especially with zero-shot learning. 
However, results on zero-shot learning have a wide variance in the published literature across tasks and languages \cite{DBLP:journals/corr/abs-2004-15001}. 
A more systematic study, controlling for the design parameters discussed above and the training and test sets is required.
Specifically, there is need for careful comparisons against translation-based baselines such as translate-test, translate-train, translate-train-all \cite{conneau2019unsupervised} across tasks and languages.\\

\noindent \textbf{mBERTologoy.} 
A large body of literature has studied what the monolingual BERT model learns, and how and where such information is stored in the model \cite{rogers2020primer}. 
One example is the analysis of the role of attention heads in encoding syntactic, semantic, or positional information \cite{voita-etal-2019-analyzing}. 
Given the similarity in architecture, such analysis may be extended to \MLLMs. 
This may help interpretability of \MLLMs~but crucially contrast multilingual models from  monolingual models, providing clues on the emergence of cross-linguality.\\
    
\noindent \textbf{Language inclusivity.} 
\MLLMs~hold promise as an `infrastructure' resource for the long list of the languages in the world.
Many of the languages are widely spoken but not sufficiently focused enough in research and development \cite{joshi2020state}.
Towards this end, \MLLMs~must become more inclusive scaling up to 1000s of languages.
This may require model innovations such as moving beyond language-specific adapters. 
Crucially, it also requires the availability of inclusive benchmarks in variety of tasks and languages.
Without such benchmark datasets, the research community has little incentive to train and evaluate \MLLMs~targeting the long list of the world's languages. 
We see this as an important next phase in the development of \MLLMs.\\

\noindent \textbf{Efficient models.} 
\MLLMs~represent some of the largest models that are being trained today. 
However, running inference on such large models is often not possible on edge devices and increasingly expensive on cloud devices. 
One important research direction is to downsize these large models without affecting accuracy.
Several standard methods such as pruning, quantization, factorization, distillation, and architecture search have been used on monolingual models \cite{tay2020efficient}. 
These methods need to be explored for \MLLMs~while ensuring that the generality of \MLLMs~ across languages is retained. \\

\noindent \textbf{Robust models.}
\MLLMs~supporting multiple languages need to be extensively evaluated for any encoded biases and their ability to generalize.
One direction of research is to build extensive diagnostic and evaluation suites such as MultiChecklist proposed in \citet{DBLP:journals/corr/abs-2104-07412}.
Evaluation frameworks such as Explainaboard \cite{liu2021explainaboard, fu2020interpretable} need to also be developed for a range of tasks and languages to identify the nature of errors made by multilingual models.
It is also important to extend the analysis of bias in deep NLP models to multilingual systems \cite{blodgett2020language}.

\section{Conclusion}
We reviewed existing literature on \MLLMs~covering a wide range of sub-areas of research. In particular, we surveyed papers on building better and bigger \MLLMs~using different training objectives and different resources (monolingual data, parallel data, back-translated data, etc). We also reviewed existing benchmarks for evaluating \MLLMs~and covered several studies which use these benchmarks to assess the factors which contribute to the performance of \MLLMs~in a (i) zero-shot cross-lingual transfer setup (ii) monolingual setup and (iii) bilingual setup. Given the surprisingly good performance of \MLLMs~in several zero-shot transfer learning setups we also reviewed existing works which investigate whether these models learn any universal patterns across languages. Lastly, we reviewed studies on improving the limited capacity of \MLLMs~and extending them to new languages. Based on our review, we recommend that future research on \MLLMs~should focus on (i) controlled ablation studies involving a broader set of parameters (ii) comprehensive evaluation of the zero-shot performance of \MLLMs~across a wider set of tasks and languages (iii) understanding the patterns learn by attention heads and other components in \MLLMs~(iv) including more languages in pretraining and evaluation and (v) building efficient and robust \MLLMs~using better evaluation frameworks (e.g., Explainaboard).

\section*{Acknowledgements}
We would like to thank the Robert Bosch Center for Data Science and Artificial Intelligence for supporting Sumanth and Gowtham through their Post Baccalaureate Fellowship Program. We also thank the EkStep Foundation for their generous grant to support this work.

\bibliography{anthology,custom}
\bibliographystyle{acl_natbib}

\end{document}